\documentclass[12pt]{article}
\DeclareUnicodeCharacter{0301}{\'{e}}
\usepackage{amssymb}
\usepackage{amsmath}
\usepackage{amsthm}
\usepackage{color}
\usepackage{comment}
\usepackage{enumitem}		
\usepackage{geometry}		
\usepackage{graphicx}
\usepackage{authblk}
\usepackage{amssymb}
\usepackage{listings}
\usepackage{xcolor}
\usepackage{caption}
\usepackage{color}
\usepackage{algorithm}
\usepackage{algpseudocode}

\usepackage{subcaption}

\makeatletter
	\@addtoreset{equation}{section}
\makeatother

\let\originalleft\left
\let\originalright\right
\renewcommand{\left}{\mathopen{}\mathclose\bgroup\originalleft}
\renewcommand{\right}{\aftergroup\egroup\originalright}

\newlist{romanlist}{enumerate}{3}
\setlist[romanlist]{label=\roman*),ref=(\roman*)}

\begin{document}

\newcommand{\cF}{\mathcal{F}}
\newcommand{\cP}{\mathcal{P}}
\newcommand{\cR}{\mathcal{R}}
\newcommand{\cS}{\mathcal{S}}
\newcommand{\cT}{\mathcal{T}}
\newcommand{\ee}{\varepsilon}
\newcommand{\rD}{{\rm D}}
\newcommand{\re}{{\rm e}}

\newtheorem{theorem}{Theorem}[section]
\newtheorem{corollary}[theorem]{Corollary}
\newtheorem{lemma}[theorem]{Lemma}
\newtheorem{proposition}[theorem]{Proposition}

\theoremstyle{definition}
\newtheorem{definition}{Definition}[section]


\title{
Comparative Analysis of Predicting Subsequent Steps in Hénon Map
}

\author[1,*]{Vismaya V S}
\author[1]{Alok Hareendran}
\author[1]{Bharath V Nair}
\author[1]{Sishu Shankar Muni}
\author[2]{Martin Lellep}

\affil[1]{School of Digital Sciences, Digital University Kerala, Thiruvananthapuram, PIN 695317, Kerala, India}

\affil[2]{SUPA, School of Physics and Astronomy, The University of Edinburgh, James Clerk Maxwell Building,
Peter Guthrie Tait Road, Edinburgh EH9 3FD, UK}

\maketitle


\begin{abstract}
This paper explores the prediction of subsequent steps in Hénon Map using various machine learning techniques. The Hénon map, well known for its chaotic behaviour, finds applications in various fields including cryptography, image encryption, and pattern recognition. Machine learning methods, particularly deep learning, are increasingly essential for understanding and predicting chaotic phenomena. This study evaluates the performance of different machine learning models including Random Forest, Recurrent Neural Network (RNN), Long Short-Term Memory (LSTM) networks, Support Vector Machines (SVM), and FeedForward Neural Networks (FNN) in predicting the evolution of the Hénon map. Results indicate that LSTM network demonstrate superior predictive accuracy, particularly in extreme event prediction. Furthermore, a comparison between LSTM and FNN models reveals the LSTM's advantage, especially for longer prediction horizons and larger datasets. This research underscores the significance of machine learning in elucidating chaotic dynamics and highlights the importance of model selection and dataset size in forecasting subsequent steps in chaotic systems.
\end{abstract}
\section{Introduction}

\label{sec:intro}
Chaos \cite{Jrgensen2008} refers to systems that seem random but are governed by deterministic laws. Even small changes in initial conditions can lead to vastly different outcomes over time, known as the "butterfly effect" \cite{RouvasNicolis2009}. Predicting chaotic systems is difficult due to their extreme sensitivity to initial conditions, making long-term predictions impractical. These systems often involve complex behaviours, nonlinear dynamics \cite{Toker2020}, and interactions between variables, defying simple analytical solutions.Though it has limitations in terms of forecast accuracy, chaos theory offers insights into a variety of domains and statistical regularities to examine despite its obstacles.

A mathematical system called the Hénon Map \cite{Hnon1976} describes the behavior of a two-dimensional discrete dynamical device that is nonlinear. It is named after the French mathematician and astronomer.It was initially put forth by Michel Hénon in 1976 as a simplified explanation for how stars move within galaxies.Renowned for its chaotic behavior, the Hénon Map \cite{Marotto1979} finds a huge variety of applications in domains like information protection \cite{Sukirman2014}, cryptography \cite{Ibrahim2020}, picture encryption \cite{Mishra2018}, random variety technology \cite{Ghayad2019}, sample reputation \cite{Zhang2017}, and steganography \cite{Thenmozhi2013}.Sensitive dependency on preliminary situations and nonlinear dynamics are explained by way of this foundational example within the look at of chaotic structures. It enables create pseudo-random numbers in cryptography and secure communication protocols in picture encryption, it jumbles records to save you unwanted access. Moreover, the unpredictable traits of the map lead them to beneficial for steganography a way that hides personal statistics in non-mystery information—and pattern reputation duties. The chaotic qualities of the Hénon Map render it fairly desirable for packages that necessitate confidentiality,unpredictability, and randomization.

Numerous phenomena, consisting of as bifurcations \cite{Zhou2013}, transient chaos \cite{Lai2011}, touchy dependency
on preliminary situations \cite{Lai1994}, atypical attractors \cite{Plykin1995}, and transient bursts \cite{Zich2020}, are as a result of the chaotic conduct of the H eon map. These are but a few of the striking occurrences that the map illustrates. While temporary chaos is the condition of chaos prior to periodicity or the convergence of the alternate chaotic regime, bifurcations \cite{Bury2023} illustrate how the quality of the system changes with changing parameters. Dependency on delicate starting circumstances highlights the
system's susceptibility to even little changes in the initial conditions, which might have unanticipated consequences.Strange attractors represent complex geometric patterns governing long-term behavior, whereas transient bursts represent random, rapid changes in system variables. Understanding nonlinear phenomena and chaos theory requires an understanding of the intricate dynamics of nonlinear systems, which is largely provided by these extreme events.

Machine learning techniques \cite{Dickinson2021} are becoming increasingly essential for solving intricate algorithms across diverse domains like scientific computations, climate forecasting \cite{Bochenek2022}, and medical diagnosis \cite{Erickson2017}. These methods rely heavily on extensive datasets to effectively train models. However, analysing vast datasets, especially those involving time series data, presents notable challenges. When precise mathematical rules governing a system are unknown, machine learning, particularly deep learning \cite{Schulz2012}, becomes indispensable. Nevertheless, deep learning often demands substantial data for optimal training, which might not always be accessible due to system complexity or experimental constraints. Furthermore, the dynamic nature of real-world phenomena often requires continuous adaptation and refinement of machine learning models to maintain their relevance and accuracy over time. This iterative process of model enhancement highlights the ongoing necessity for robust methodologies capable of handling evolving datasets and complex problem domains. Moreover, the interdisciplinary nature of contemporary research emphasises the significance of collaboration among experts from various fields, pooling their domain knowledge and machine learning expertise to address multifaceted challenges. In this regard, the synergy between domain-specific insights and machine learning capabilities plays a pivotal role in uncovering novel perspectives and propelling innovation across diverse disciplines.\cite{Taye2023}.

Artificial Neural Networks (ANN)\cite{Grossi2007} imitate the operational mechanisms of the human brain and are highly effective in solving nonlinear problems by adjusting their internal structure, the arrangement and organization of interconnected nodes, or neurons, within the network to achieve specific goals. They excel particularly in tasks such as pattern recognition and regression. The Hénon map produces complex behaviour from simple equations and is valuable for studying chaotic phenomena. Artificial Neural Networks versatility goes beyond traditional problem-solving domains, areas where algorithms and computational techniques have been historically applied to solve specific types of problems extending to applications in image and speech recognition \cite{Lim}, natural language processing \cite{Chowdhary2020}, and autonomous decision-making systems \cite{Chen2017}. Their capacity to learn from extensive data and adapt to changing environments makes them powerful tools for addressing the intricacies of modern computational challenges. The exploration of chaotic phenomena using the Hénon map demonstrates the broad applicability of neural network approaches, offering insights into the underlying dynamics of seemingly unpredictable systems and contributing to advancements in chaos theory and dynamical systems analysis.

The present paper explores predicting subsequent steps in the Hénon map using machine learning techniques. Various methods such as Random Forests \cite{Breiman2001}, Recurrent Neural Networks (RNN) \cite{Sherstinsky2020}, Long Short-Term Memory (LSTM) networks \cite{Sherstinsky2020}, and Support Vector Machines (SVM) \cite{Zhang2020} are considered.The study demonstrates that adjusting the topology of machine learning models can significantly enhance performance, indicating that accuracy is not solely dependent on the volume of training data. By employing different methodologies and inducing topological changes, the models' predictive capabilities are substantially improved.

The novelty of the paper includes the following:
\begin{itemize}
    \item \textcolor{black}{The paper initiates an inquiry into the effectiveness of machine learning algorithms for predictive tasks, with the objective of assessing their effectiveness in predictive modeling.}
    \item \textcolor{black}{Through meticulous analysis and evaluation, the study systematically compares and contrasts various techniques within the machine learning domain.} 
    \item \textcolor{black}{The primary objective is to identify the most effective machine learning approach among those examined.}
\end{itemize}

The paper is organised as follows: Section 2 explores into the Hénon Map, offering a detailed explanation of its mathematical underpinnings and relevance to our analysis. Following this, in Section 3, we explore predictions using machine learning methods, discussing the techniques employed and their application to our research objectives. Section 4 presents the results obtained from our analyses, elucidating key findings and insights derived from the data. We then proceed to Section 5, where we offer a comprehensive conclusion summarizing our study's outcomes and implications.

\section{Hénon Map}
\begin{figure}[!hbtp]
    \centering
    \includegraphics[width=0.8\linewidth]{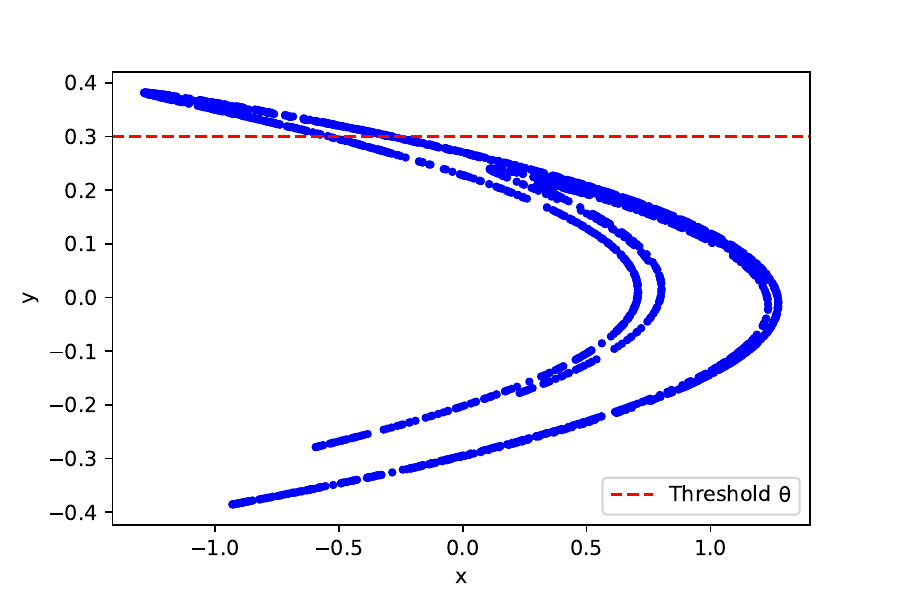}
    \caption{The H\'{e}non-map chaotic attractor shown as blue dots and dashed red lines represent the threshold $\theta$}
    \label{Fig.1}
\end{figure}

\begin{figure}[!hbtp]
    \centering
    \includegraphics[width=0.77\linewidth]{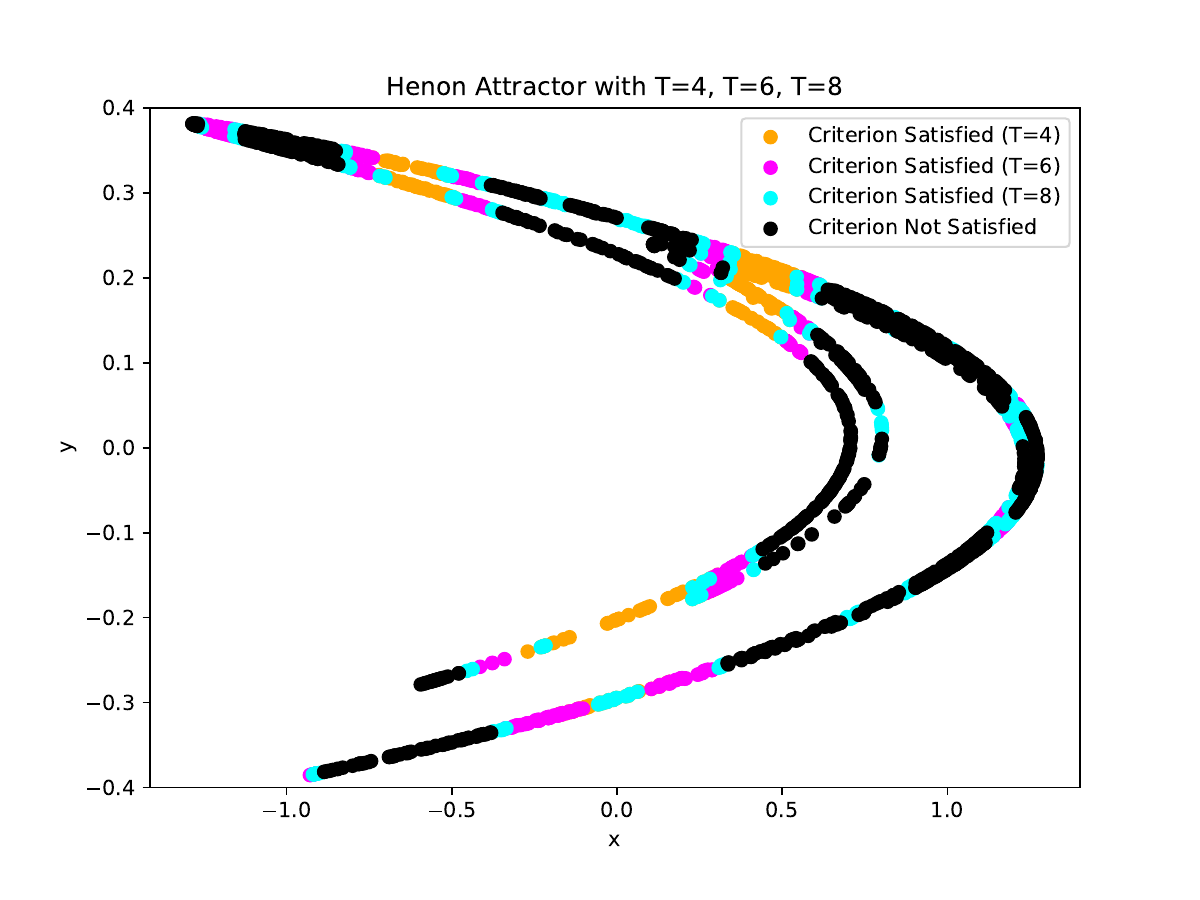}
    \caption{H\'{e}non Attractor with time steps T = 4, 6, and 8 where black: Criterion not satisfied; Orange, Magenta, Cyan: Criterion satisfied (T = 4, 6, 8)}
    \label{Fig:2}
\end{figure}
The Hénon map \cite{Lellep2020} gives a vital paradigm for comprehending chaotic action within 
basic mathematical frameworks. This map is based on a fixed iterative equations that describe
the evolution of the system state from one step to the following. More precisely, the behaviour of
the Hénon map is described by equations that describe the machine’s time-various coordinates.
These formulae offer a compact yet effective system for analyzing complicated phenomena consisting of chaos
and encapsulate the essence of the system dynamic. The Hénon map is defined by the following iterative equations:
\begin{equation}
\begin{aligned}
x_{n+1} &= 1 - ax_n^2 + y_n, \\
y_{n+1} &= bx_n
\end{aligned}
\label{eq:1}
\end{equation},
where \(x_n\),\(y_n\) are the coordinates of a point in the plane at time step \(n\), and \(a\) and \(b\) are parameters that determine the behaviour of the system. Here \(a = 1.4\) and \(b = 0.3\) and a threshold $\theta = 0.3$  has been used as shown in the Fig.\ref{Fig.1}

Now we used a criterion,
\begin{equation}
\begin{aligned}
y_{n+T} \geq \theta
\end{aligned}
\label{eq:2}
\end{equation}
where we have fixed $\theta = 0.3$.

\textcolor{black}{In the given plot Fig.\ref{Fig:2}, the variable T represents the number of future steps to be considered when checking whether the criterion is satisfied for a given point on the Hénon Attractor. Specifically, the criterion eqn:\eqref{eq:2} is satisfied if the y-coordinate of the point at time t+T is greater than or equal to a specified threshold value $\theta$.The threshold value $\theta$ is a fixed constant that determines the level at which the criterion is considered to be satisfied.
In the given plot, the threshold value is the same for all three criteria ($\theta = 0.3$).The plot shows the trajectory of the Hénon Attractor, along with the points that satisfy the criteria for T = 4, 6, and 8. The orange dots correspond to points that satisfy the criterion for T = 4, the magenta dots correspond to points that satisfy the criterion for T = 6, and the cyan dots correspond to points that satisfy the criterion for T = 8. The black dots correspond to points that do not satisfy any of the criteria.Overall, the plot provides a visual representation of the behavior of the Hénon Attractor under the criteria and threshold value.}

\section{Predictions using Machine Learning Methods}
We use different machine learning models for the prediction of \textcolor{black}{ subsequent steps} in the Hénon Map.
In the paper "Using Machine Learning to predict extreme events in the Hénon map" \cite{Lellep2020} utilizes a feedforward neural network \cite{Bebis1994}. However, it has not yet conducted a comparative analysis of different machine learning techniques. Such a comparison is crucial for determining the most appropriate and effective approach for the task. Here, we are using Random Forest \cite{Pradeepkumar2016},RNN \cite{Bertschinger2004},LSTM \cite{Yanan2020}, and SVM \cite{Ye2005} for the prediction.

\begin{figure}[!hbtp]
    \centering
    \includegraphics[width=1\linewidth]{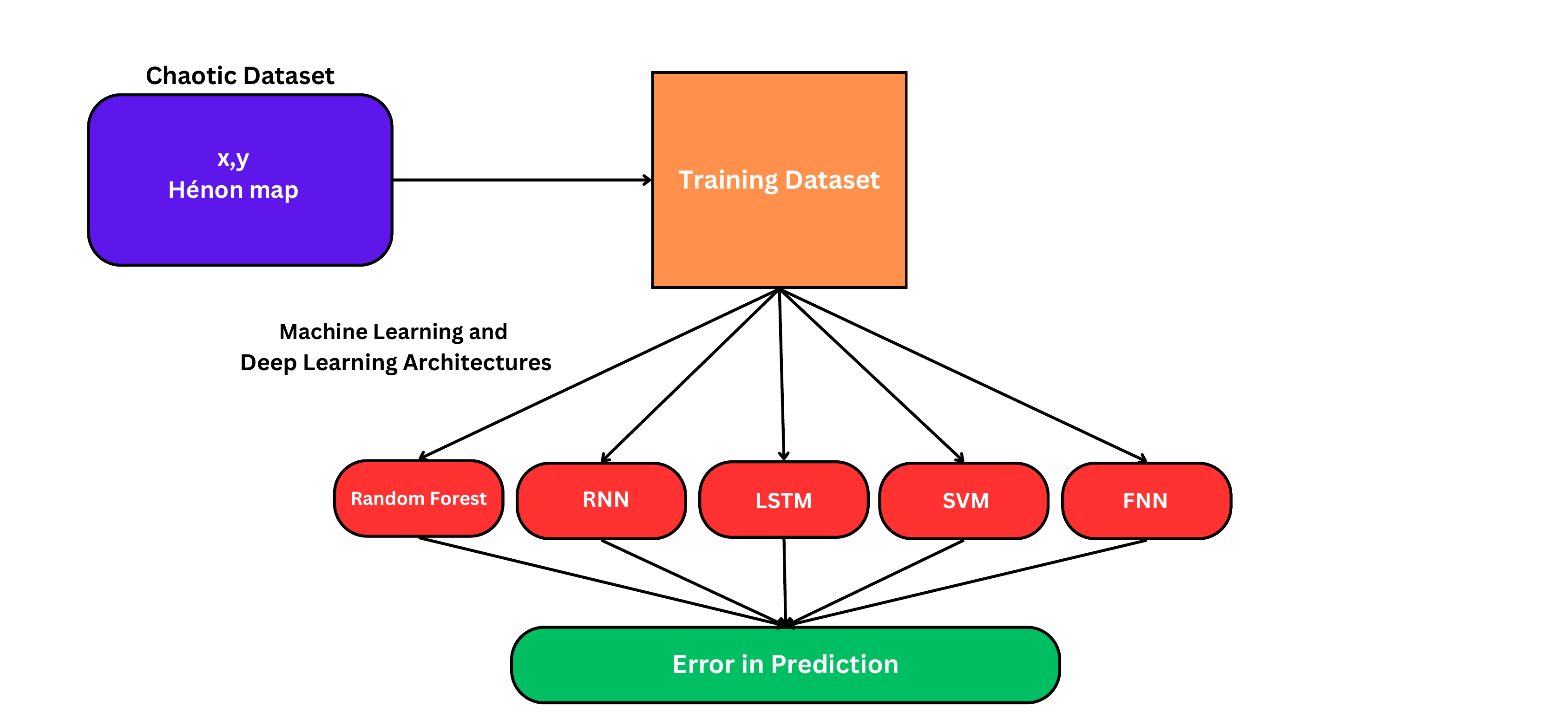}
    \caption{Schematic diagram showing the process of generating training data, training the prediction model, and
finally using the trained model for prediction}
    \label{Fig:3}
\end{figure}

The Hénon map data is generated by iterating the Hénon map equations eqn:\eqref{eq:1} for 10,000 iterations. For the equation we have taken \(a = 1.4\) and \(b = 0.3\). Then \textcolor{black}{only last 20\% of the data is taken because focusing on the last 20\% of the data helps to highlight the long-term behavior of the system, as initial conditions can heavily influence short-term dynamics. By excluding the initial transient behavior, the plot better captures the attractor or stable pattern of the system.Then the data is splitted into training and testing sets, with 80\% and 20\% respectively for the evaluation.}The following models are used for our prediction
\subsection{Random Forest}
Random Forest (RF) regression model has been implemented and the training data, consisting of sequential pairs of coordinates representing the  Hénon map, is appropriately formatted for input to the RF regressor. The regressor, configured with 100 decision trees, is then fitted to the training data, leveraging the ensemble learning capabilities of Random Forest to capture the complex and nonlinear patterns inherent in the Hénon map dynamics.

Subsequently, the trained RF regressor is employed to predict the subsequent steps of the Hénon map and MSE is calculated.

\subsection{Reccurent Neural Network(RNN)}
A Recurrent Neural Network (RNN) architecture has been implemented. First, a SimpleRNN layer with 10 units and a hyperbolic tangent activation function is added in the model's sequential construction.Within this architecture, the RNN component that is responsible for processing sequential enter is known as the SimpleRNN layer. In order to learn the model to recognise temporal
dependencies in the data, it keeps tune of information about in earlier time steps within the sequence. This layer's internal recurrent connections make it especially useful for capturing short-term dependence. However, the Dense layer is a typical fully connected neural network layer that comes after the SimpleRNN layer. Each neuron in the layer above is linked to every other neuron, which is how it gets identified.The traits that the SimpleRNN layer has learnt on this model need to be formatted efficiently for the very last prediction task with the aid of way of the Dense layer. A sequence of weighted connections and activation features are used to carry out this change, which lets in the model to recognize complex patterns and correlations within the information.In order to fine-tune
the parameters of the model for precise prediction, it is trained using the Adam optimizer with a learning rate of 0.001 and mean squared error as the loss characteristic.
The training technique involes 50 epochs for the model, in which an epoch is one full run through the training dataset. The model dealt with 32 samples of information in each training iteration due to the fact that a batch size of 32 was used. In particular, when running with big datasets, this batch-wise training strategy facilitates to optimise memory utilisation and computing performance. The model is used to forecast how the H\'{e}non map in the trying out set will change once it has been trained.The mean squared error  
(MSE) among the expected and actual is computed to assess the prediction’s accuracy.

\subsection{Long Short-Term Memory (LSTM)}
A neural community with long Short-Term Memory (LSTM) has been implemented. The successive
coordinate pairs that make up the training data are definitely changed to meet the LSTM model’s
input specs. A sequential model is used to build the model architecture, which includes
a 10-unit LSTM layer and a hyperbolic tangent activation function. The sequential model is a linear stack of layers with precisely one input tensor and one output tensor for every layer. Building neural networks is made simple with the aid of this easy structure, which allows layer addition in a sequential way. In this example, the layers of the LSTM model are organized inside the Sequential model.

Recurrent neural network (RNN) layers which includes the LSTM layer were created expressly to
deal with the vanishing gradient problem that conventional RNNs have. By preserving a memory state
and step by step updating it with alternatives, it can realize long-term dependencies in sequential data.
A sequence of memory cells with input, forget, and output gates make up the LSTM layer. The model is able to hold considerable data over extented durations while discarding inappropriate in-
formation because these gates modify the flow of data into and out of the cells. This design decision is a reflection of knowledge about the temporal relationships seen in chaotic structures, such
the Hénon map. The two dimensions of the Hénon map are matched by a second Dense layer with
two units which makes an intensive prediction feasible. The training process is successfully guided by the mean squaered error loss function, that is hired together with the Adam optimiser
with a learning rate of 0.001 to construct the model.
There were 50 epochs inside the training procedure for the model, in which one epoch is one full
run via the training dataset. 32 data samples have been processed with the aid of the model in each training
iteration for the reason that a batch size of 32 was employed. When operating with big datasets, specially,
this batch-wise training method helps optimise memory utilisation and computing overall performance. The Hénon map's development in the testing set is then predicted using the trained LSTM model.  The predictive accuracy is quantified through the calculation of the mean squared error (MSE).
\subsection{Support Vector Machine (SVM)}
The regression model used is the Support Vector Machine (SVM) model. We have utilised a window size of 5—the number of consecutive data points used as input to the model—to prepare the data for training. The input-output pairs that make up the dataset's structure are pairs of coordinates representing the next coordinate in the development of the Henon map, and each input is made up of a series of coordinates. Data is presented according to this window size for both training and testing. Both sets of data had their data flattened to comply with SVM standards. With an emphasis on predicting the x-coordinate, we have trained and instantiated the SVM model using a linear kernel for simplicity.

After training, we've used the fitted SVM model to predict the H\'{e}non map evolution in the testing set. Then, we've calculated the mean squared error (MSE) to measure the difference between the predicted and actual values, providing a quantitative assessment of the model's predictive accuracy.

For the above models, we used the activation function \cite{Rasamoelina2020} which is a crucial component of artificial neural networks. An activation function in the context of a neural network is a mathematical function used to transform the summed weighted input from the neuron into an output that is then passed on to the next layer. Activation functions help introduce non-linearity into the neural network, allowing it to learn and model complex patterns and relationships in the data.

There are several types of activation functions, each with its own unique properties and use cases. Some of the most commonly used activation functions include Step Function, Linear Function, Sigmoid function \cite{Pratiwi2020}, Tanh function and ReLU (Rectified Linear Unit) function .
\begin{figure}[!hbtp]
    \centering
    \begin{minipage}[b]{0.33\textwidth}
        \centering
        \includegraphics[width=1\textwidth]{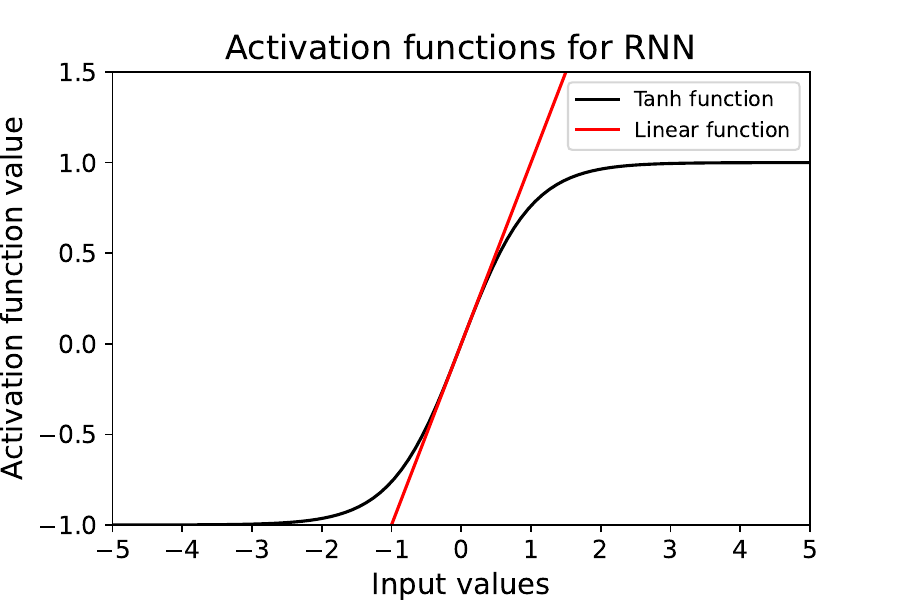}
        \caption*{(a)}
    \end{minipage}
    \hfill
    \begin{minipage}[b]{0.33\textwidth}
        \centering
        \includegraphics[width=1\textwidth]{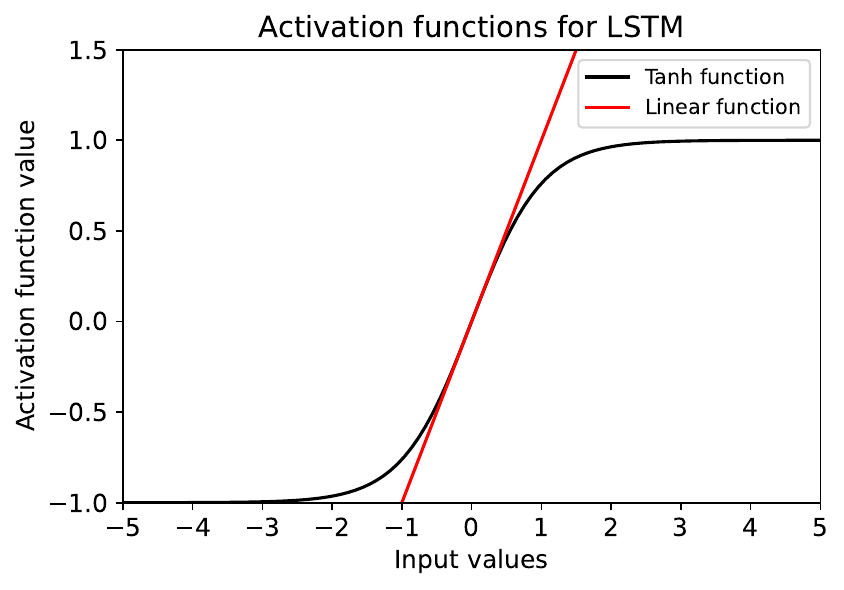}
        \caption*{(b)}
    \end{minipage}
    \hfill
    \begin{minipage}[b]{0.33\textwidth}
        \centering
        \includegraphics[width=1\textwidth]{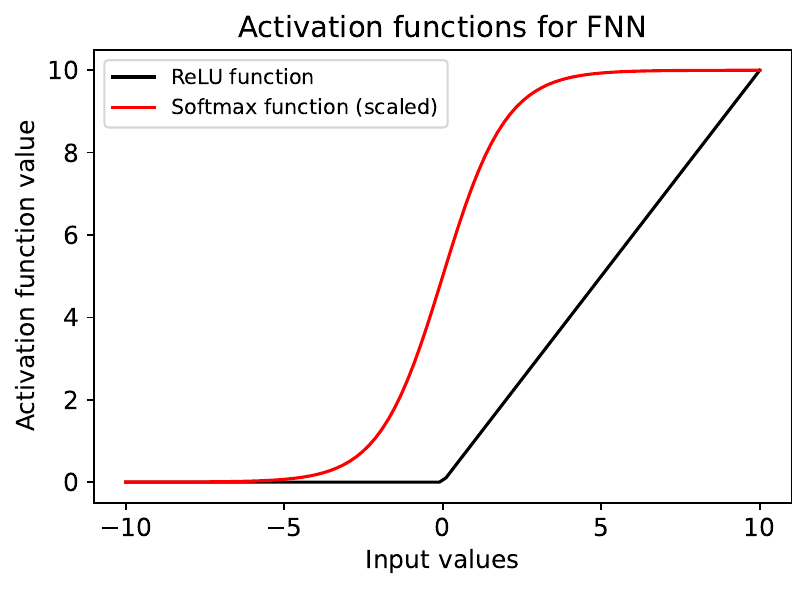}
        \caption*{(c)}
    \end{minipage}
    \caption{Activation functions for hidden and output layers of different architectures;
    (a) Activation functions for RNN, (b) Activation functions for LSTM, (c) Activation functions for FNN}
    \label{Fig.4}
\end{figure}

Following are the activation functions we used for different architectures:

\begin{itemize}
    \item {\textit{Activation Function for RNN}} 
\end{itemize}
A plot of the Tanh and Linear activation functions,is presented in Figure \ref{Fig.4} (a) to show how they are distinct from one another. For the two functions, the Tanh function (black line) and the Linear function (red line), the figure contrasts the input values (x-axis) with the respective activation function values (y-axis). The Tanh function is sigmoidal, as the figure demonstrates, with output values ranging from -1 to 1.The differences between the two activation functions' characteristics and how well they fit various RNN model types are demonstrated by this evaluation.

\begin{itemize}
    \item{\textit{Activation Function for LSTM}}
\end{itemize}
The activation functions of LSTM networks have a key role for figuring out how the network’s
input and output values are formed. The Tanh characteristic, which transfers the input values to a
range among -1 and 1, is a sigmoidal function, as visible in figure \ref{Fig.4} (b). It is standard exercise
to hire this function to manipulate records drift between LSTM network units and to inject
non-linearity into the community. By comparison, the linear characteristic outputs the enter values immediately
out of the field with none change. The transfer of facts through the network is controlled by means of some
LSTM units using this function.

The figure suggests the input values at the x-axis and the corresponding activation function values on the y-axis. The Tanh characteristic is plotted in black and the Linear feature is plotted in red As can be seen, the Tanh characteristic is sigmoidal in shape, with output values ranging among -1 and 1, while the Linear feature is a straight line, with output values identical to the enter values. By the use of those activation features in LSTM networks, researchers and practitioners can design networks which can procedure sequential facts and examine complex patterns over the years. The preference of activation functions could have a full-size effect on the performance of the network, and therefore it's miles crucial to understand their homes and limitations.
\begin{itemize}
    \item{\textit{Activation Function for FNN}}
\end{itemize}

Softmax Fig.\ref{Fig.4} (c) and Rectified Linear Unit (ReLU) \cite{agarap} are two activation functions for feedforward neural networks (FNN) that are visualised. In order to provide non-linearity and enable the network to learn intricate patterns, these activation functions are frequently employed in neural networks. A common and straightforward activation function is called Rectified Linear Unit, or ReLU. It assigns a value of zero to every negative input value and a value of themselves to every positive value. This indicates that there is always a non-negative result from the ReLU function. In the output layer of a neural network for multi-class classification issues, another popular activation function is called Softmax.It maps the input values into a probability distribution over a set of classes. The output values of the softmax function sum up to 1, which means that they form a valid probability distribution.

The plot shows that the ReLU function is piecewise linear, with a slope of 1 for positive input values and 0 for negative input values. The Softmax function, on the other hand, is a smooth curve that maps the input values into a probability distribution.

\begin{figure}[!hbtp]
    \centering
    \includegraphics[width=1\linewidth]{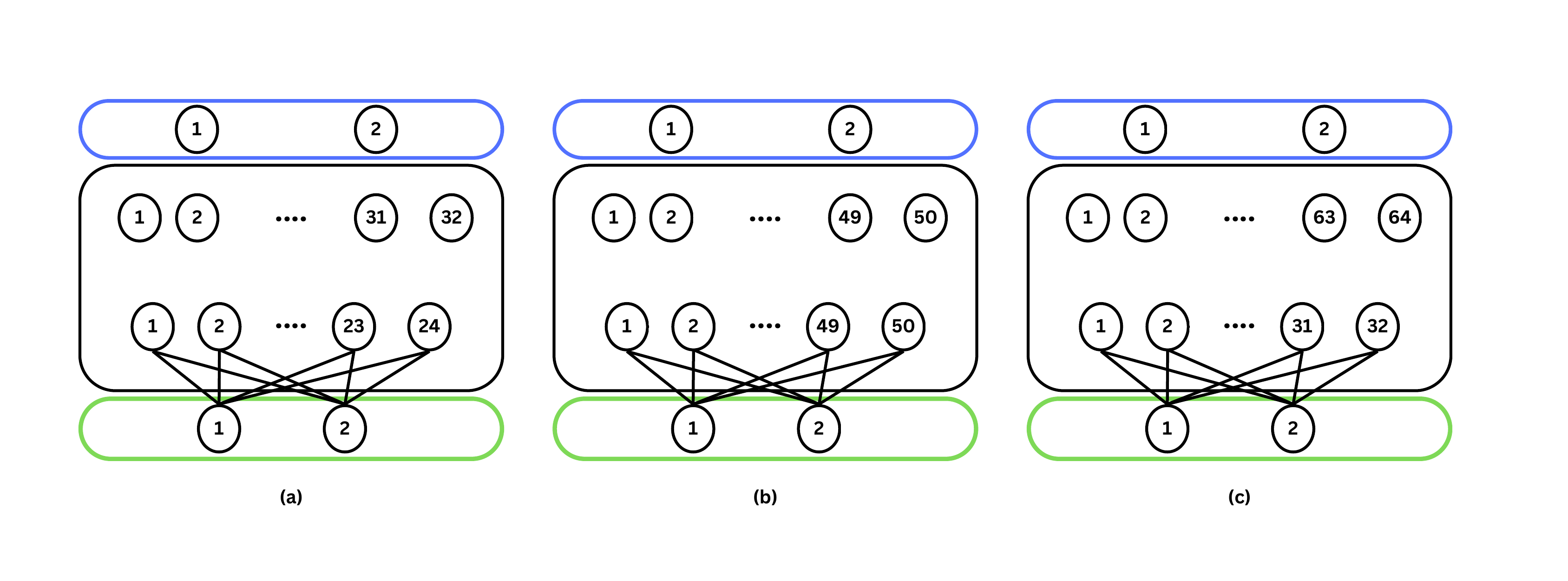}
    \caption{Topology of the neural network. There are input nodes (blue box), hidden layers (black box), and output neurons (green box), where N is the number of previous time steps.
    (a) Recuuent Neural Network (RNN),(b) Long Short-Term Memory (LSTM),(c) Feedforward Neural Networks (FNN)}
    \label{Fig:5}
\end{figure}

From the above architectures, as shown in Fig. \ref{Fig:5} all of the connections inside the network are connected. Every neuron in the hidden and input layers is linked to every other neuron in the layer below it. To prevent having too many weights and creating too much visual load, just the weights from the final hidden layer to the output layer are shown. The numbers indicate each layer's neurons. In the input layer of the RNN, LSTM and FNN models, there are three neurons, each capturing distinct features or dimensions of the input data. This foundational layer serves as the entry point for information processing in the neural network, where input data is received and processed at each time step, facilitating sequential analysis such as time series forecasting or natural language processing. Regardless of the model type, the input layer's neurons play a crucial role in capturing the essential characteristics of the input data.

In order to extract and represent underlying patterns inside the sequential facts, every model makes use of
a one of a kind shape in its hidden layers. A usual of 32 neurons make up the number one layer and
24 neurons make up the second one recurrent hidden layer of the RNN structure. By keeping
a temporal memory and capturing dependencies at some point of successive inputs, the network is capable of
maintain those recurrent connections. LSTM, but, makes use of two hidden layers with fifty
reminiscence-gated neurons each. By solving the vanishing gradient issue and making it easier to identify long-term relationships, these gates enable LSTMs to precisely keep or discard data over lengthy periods. First, there are 64 neurons in the first layer and 32 neurons in the second completely linked hidden layer of the FNN. 

Finally, the output layer of all three models consists of two neurons, serving as the endpoint for prediction tasks. These neurons receive inputs from the final hidden layer and generate predictions based on learned patterns in the data. Whether forecasting future time steps or making predictions on structured data, the output layer synthesizes the network's learned representations into actionable insights.

\section{Results}
In our study, we aimed to predict subsequent steps in the Hénon Map using different machine learning models.After the training,MSE of each model is obtained.
For the models Random Forest, RNN, LSTM, SVM obtained a mean squared error (MSE) of 0.5590726121211392, 0.5651870153693168, 2.0027390680736243e-06 and 0.4274929286882515 respectively.

We also used the model Feedforward Neural Network (FNN) using \textit{TensorFlow} and \textit{Keras} to predict the Hénon map's evolution. The FNN architecture consists of two hidden layers with Rectified Linear Unit (ReLU) activation functions and an output layer with softmax activation. The model is trained with the Adam optimizer \cite{Zhang2018} and mean squared error loss function. Following training, the model predicts the Hénon map's evolution on the testing set. The obtained Mean Squared Error (MSE) value of 0.6486854565019292 quantifies the model's predictive accuracy. Then we have carried out a comparative study of MSE of different models.

\begin{figure}[!hbtp]
    \centering
    \includegraphics[width=0.8\linewidth]{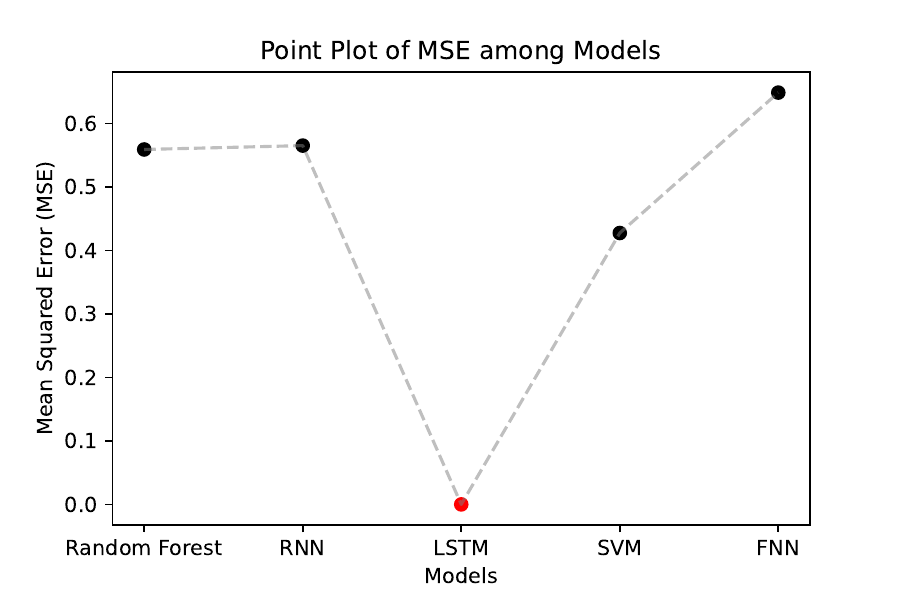}
    \caption{A point plot of Mean Squared Error (MSE) for five models: Random Forest, RNN, LSTM, SVM, and FNN.}
    \label{Fig:6}
\end{figure}

\begin{figure}[!hbtp]
    \centering
    \includegraphics[width=0.8\linewidth]{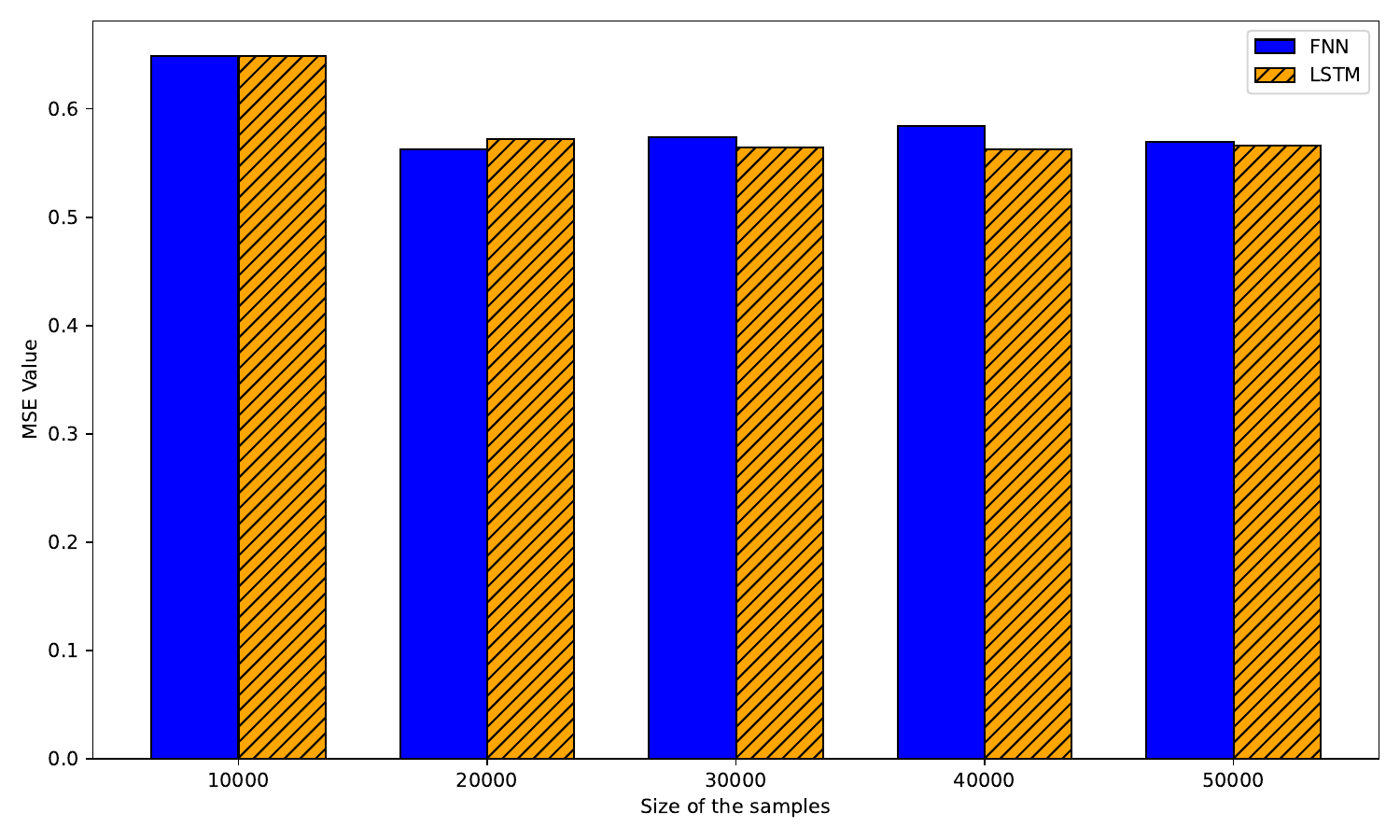}
    \caption{Bar plot showing the performance of FNN and LSTM models in terms of Mean Squared Error (MSE) for different sample sizes}
    \label{Fig:7}
\end{figure}

The given figure Fig.\ref{Fig:6} is a point plot that compares the Mean Squared Error (MSE) values of different machine learning models. The models used are Random Forest, RNN, LSTM, SVM, and FNN. The MSE values for each model are plotted on the y-axis, while the names of the models are on the x-axis. As can be seen in figure \ref{Fig:6}, the LSTM model performs best in forecasting the evolution of the Hénon Map since it has the lowest MSE value, which is around 2.00e-06.With respect to MSE values, the Random Forest model is values at 0.559, the RNN model at 0.565, the
SVM model at 0.427, and the FNN model at 0.649.
The MSE values between the diverse models can be quick and without problems as compared by way of using the
point plot. The LSTM is the most accurate in forecasting the evolution of the Hénon map
on this example, as visible by using its lowest MSE value.The MSE values of the Random Forest and RNN are kind of comparable, however the SVM model’s MSE value is especially lower than that
of the FNN.
In end, the point plot affords an smooth-to-recognize visualisation of the MSE values for
several machine-learning models which can be employed to forecast the Hénon map’s subsequent steps.
The most accurate model is the LSTM model, which has the low MSE value; the least correct model is the FNN model, which has the very best MSE value. The quality model to use for forecasting
the H ́enon Map’s subsequent steps can be selected with this facts .
Eventually, we reap a bar plot Figure.\ref{Fig:7} that compares the Mean Squared Error (MSE) values of 
models,FNN and LSTM with diffrent sample sizes. The figure indicates two sets of bars: an orange
set for LSTM and a blue set for FNN. The period of the samples is proven by using the x-axis, which is going
from 10,000 to 50,000 in 10,000 increments. The difference among the actual and predicted values
is represented with the aid of the MSE value at the y-axis. The figure show that the FNN  plays
better for smaller sample numbers because it has a lower MSE value. However, the LSTM  starts offevolved to carry out better with growing pattern size, with a reduced MSE value for the best
sample length.

\begin{figure}[!hbtp]
    \centering
    \includegraphics[width=0.8\linewidth]{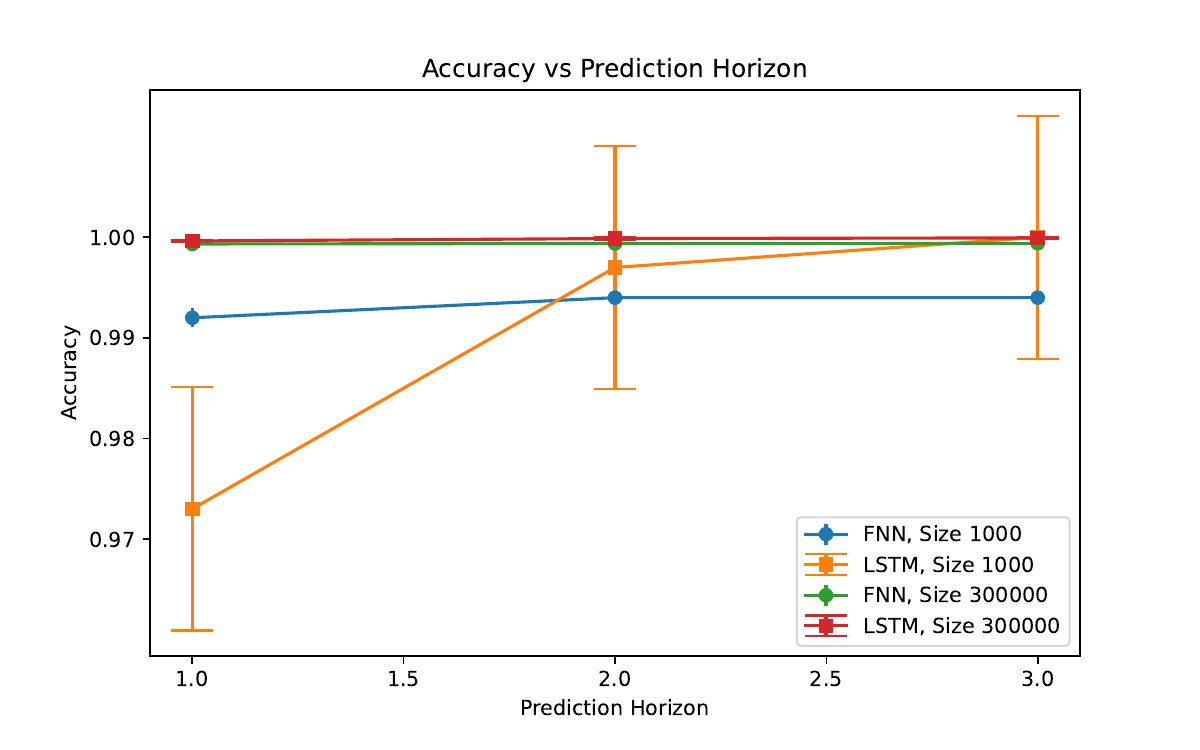}
    \caption{Comparison of FNN and LSTM accuracy across prediction horizons, datasets }
    \label{Fig:8}
\end{figure}

Accordingly, the figure indicates that, in terms of MSE performance, the FNN model would be more suited for smaller datasets, while the LSTM model might be more appropriate for bigger datasets. With the help of accuracy, we can use the LSTM and FNN models for a comparative analysis based on the data mentioned above. Figure \ref{Fig:8} compares the accuracy of two models, FNN and LSTM, at various prediction horizons. A prediction horizon is the number of iterations or future time steps that the map is used to anticipate a dynamical system's behaviour.Dataset with 1000 samples and 300,000 samples are taken for both FNN and LSTM and then trained three times with different seeds. A dataset with 1000 samples and a dataset with 300,000 samples are represented by two lines in each model in the figure. Plotting of the prediction horizon and accuracy takes place on the x- and y-axis, respectively. For every model at every forecast horizon, the error bars show the accuracy standard deviation. The LSTM model with the same dataset size is closely followed by the CNN model with the best accuracy across all prediction horizons, both with a 300,000 dataset. In both cases, the accuracy of the models drops with increasing prediction horizon; however, with larger datasets, the LSTM model continues to outperform the CNN model. 
The LSTM model outperforms the FNN model for longer prediction horizons, especially when using a larger dataset. Overall, the figure highlights the importance of selecting an appropriate prediction horizon and model for time series forecasting tasks.

\begin{figure}[!hbtp]
    \centering
    \includegraphics[width=0.8\linewidth]{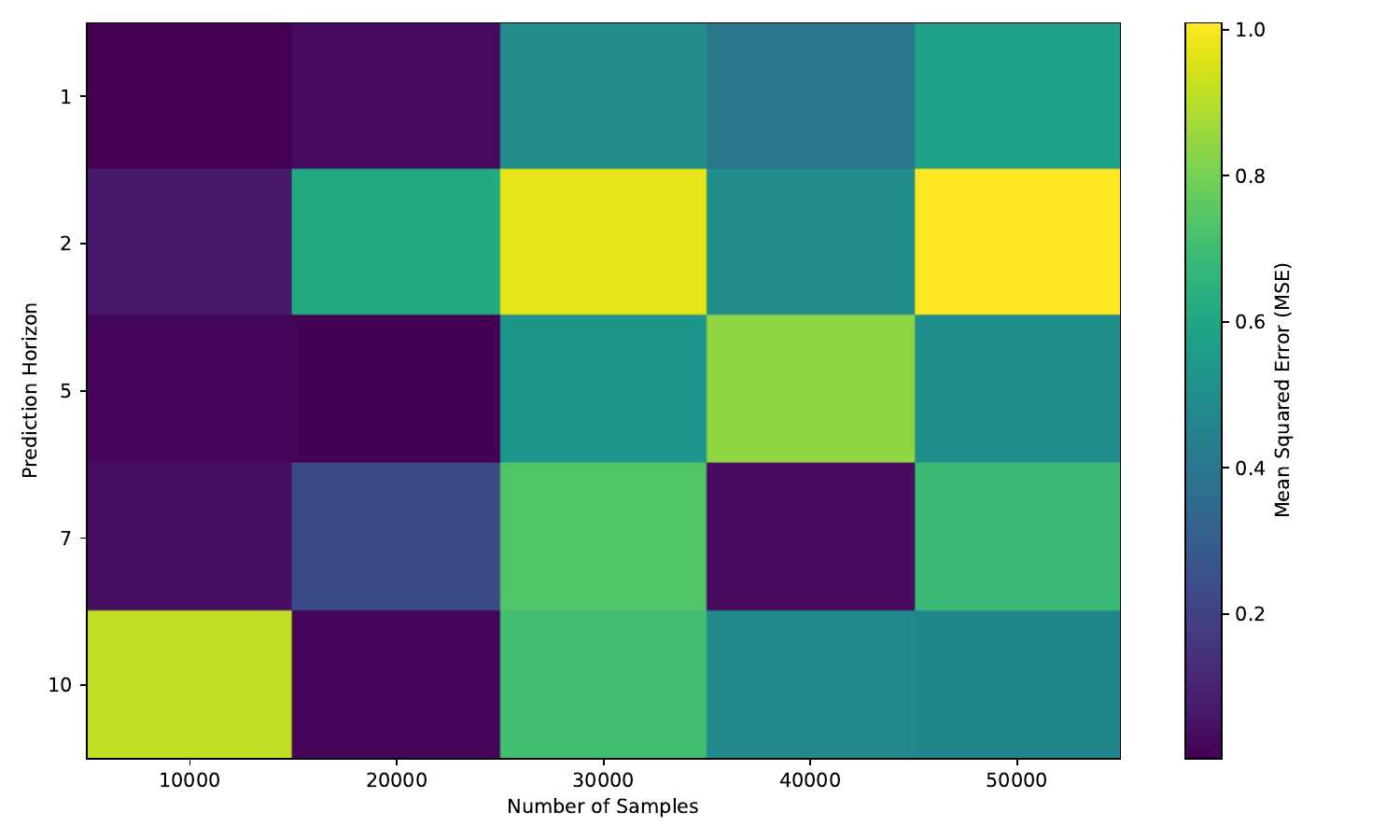}
    \caption{Heatmap of Mean Squared Error (MSE) for Predicting Next n Steps in Hénon Map Attractor with Sample Sizes of 10,000 to 50,000 and Prediction Horizons of 1 to 10}
    \label{Fig:9}
\end{figure}

Finally, we generated a heatmap Fig. \ref{Fig:9}, to visualise the Mean Squared Error (MSE) of a prediction model for different prediction horizons and sample sizes. The x-axis of the heatmap represents the number of samples used for training the model, while the y-axis represents the prediction horizon, which is the number of time steps the model tries to predict into the future.
The colour of each cell in the heatmap corresponds to the MSE value for the specific combination of sample size and prediction horizon. Darker colours indicate higher MSE values, meaning that the predictions are less accurate for those configurations. On the other hand, lighter colours represent lower MSE values, indicating more accurate predictions.
From the heatmap, it is evident that increasing the number of training samples generally leads to more accurate predictions, as seen by the lighter colours in the lower-right part of the figure. This observation is consistent with the common understanding that more training data usually improves the performance of machine learning models.

It is not as simple to determine how the prediction horizon affects the MSE, though. The MSE typically rises with the prediction horizon for smaller sample sizes; for bigger sample numbers, the link is less evident. while there is enough training data, the prediction model may be able to retain acceptable accuracy even while making predictions many steps ahead of time, as seen by the lighter-colored cells that occasionally appear in the heatmap's centre.

In summary, the produced graphic aids in the analysis of the prediction model's effectiveness for various training sample sizes and prediction horizons based on the Henon map.Although the link between the prediction horizon and MSE is less evident, the model can still predict many steps into the future with good accuracy for higher sample numbers. In the framework of time-series prediction problems, this research emphasises the significance of both training sample size and prediction horizon.

\section{Conclusion}
In order to predict subsequent steps inside the H’enon map, we used methodologies in machine learning and computational forecasting strategies. Our analysis of chaotic systems highlights their unexpected
behaviour and the problems in predicting their behaviour. To forecast subsequent steps in the
H’enon map, we made use of Random Forest, Recurrent Neural Networks
(RNN), Long Short-Term Memory (LSTM) networks, and Support Vector Machines (SVM). These
strategies have helped forecast chaotic systems and display their hidden tendencies and dynamics. Our
comparative evaluation of different methodologies highlights their strengths and boundaries in
anticipated accuracy, resilience, and overall performance. We used a FeedForward Neural
Network (FNN) , as defined in \cite{Lellep2020} and found that the FNN model outperformed others, especially when trained on a 300,000 dataset. The
predictive ability of FNN and LSTM decreased as expected whilst the prediction horizon
broadened. However, the LSTM consistantly outperformed other models throughout various dataset sizes. The LSTM network’s capability to collect long-time period relationships and temporal
dynamics in chaotic structures helped forecast results. The LSTM model is the better choice for
predicting subsequent steps in chaotic systems due to its extremely good accuracy across various prediction
horizons and dataset sizes. The Random Forest with its ensemble learning abilities
and versatility, allowed for dependable predictions in chaotic structures. RNN and LSTM networks excel
in identifying temporal patterns and modeling due to their recurrent format and
memory cells. Support Vector Machines provide a reliable framework for awaiting subsequent
steps inside the H’enon map because it recognises complex decision boundaries.

Fig.\ref{Fig:7} suggests overall performance
models among FNN and LSTM models depending on sample size, emphasizing the necessity of
selecting the ideal model for great prediction accuracy. Fig.\ref{Fig:9} shows the MSE of a prediction
model at different horizons and data samples. Lighter colorations advocate lower MSE which shows
higher performance. Larger sample sizes can retain reasonable accuracy over prolonged prediction horizons, which makes them essential for time-series forecasting endeavours using the Henon map.

Looking ahead, the adventure keeps as we embark on new frontiers of discovery and innovation. By systematically evaluating the strengths and weaknesses of each approach, we have laid the groundwork for future research and exploration in the realm of chaotic dynamics and computational forecasting. The incorporation of advanced neural network architectures, such as attention mechanisms and transformer models, represents a promising avenue for enhancing our predictive models. Investigating the effects of different parameters in the H\'{e
}non map could offer a detailed comprehension of model robustness, thereby enriching our
insights into the dynamics of chaotic systems. Further exploration of the mechanisms considered in this study can help in guiding predictions, thereby augmenting their utility in real-world applications across diverse domains such as climate modeling, finance, and beyond, with
possibilities extending beyond the mentioned fields.

\clearpage

\bibliographystyle{plain}
\bibliography{Arxiv}

\begin{thebibliography}{10}

\bibitem{agarap}
Abien~Fred Agarap.
\newblock Deep learning using rectified linear units (relu), 2018.

\bibitem{Bebis1994}
G.~Bebis and M.~Georgiopoulos.
\newblock Feed-forward neural networks.
\newblock {\em IEEE Potentials}, 13(4):27–31, October 1994.

\bibitem{Bertschinger2004}
Nils Bertschinger and Thomas Natschl\"{a}ger.
\newblock Real-time computation at the edge of chaos in recurrent neural networks.
\newblock {\em Neural Computation}, 16(7):1413–1436, July 2004.

\bibitem{Bochenek2022}
Bogdan Bochenek and Zbigniew Ustrnul.
\newblock Machine learning in weather prediction and climate analyses—applications and perspectives.
\newblock {\em Atmosphere}, 13(2):180, January 2022.

\bibitem{Breiman2001}
Leo Breiman.
\newblock Random forest.
\newblock {\em Machine Learning}, 45(1):5–32, 2001.

\bibitem{Bury2023}
Thomas~M. Bury, Daniel Dylewsky, Chris~T. Bauch, Madhur Anand, Leon Glass, Alvin Shrier, and Gil Bub.
\newblock Predicting discrete-time bifurcations with deep learning.
\newblock {\em Nature Communications}, 14(1), October 2023.

\bibitem{Chen2017}
Xue-mei Chen, Min Jin, Yi-song Miao, and Qiang Zhang.
\newblock Driving decision-making analysis of car-following for autonomous vehicle under complex urban environment.
\newblock {\em Journal of Central South University}, 24(6):1476–1482, June 2017.

\bibitem{Chowdhary2020}
K.~R. Chowdhary.
\newblock {\em Natural Language Processing}, page 603–649.
\newblock Springer India, 2020.

\bibitem{Dickinson2021}
Jennet~Elizabeth Dickinson.
\newblock {\em Machine Learning Techniques}, page 81–89.
\newblock Springer International Publishing, 2021.

\bibitem{Erickson2017}
Bradley~J. Erickson, Panagiotis Korfiatis, Zeynettin Akkus, and Timothy~L. Kline.
\newblock Machine learning for medical imaging.
\newblock {\em RadioGraphics}, 37(2):505–515, March 2017.

\bibitem{Ghayad2019}
Naeem~Howrie Ghayad and Ekhlas~Abbas Albahrani.
\newblock A combination of two-dimensional hénon map and two-dimensional rational map as key number generator.
\newblock In {\em 2019 First International Conference of Computer and Applied Sciences (CAS)}. IEEE, December 2019.

\bibitem{Grossi2007}
Enzo Grossi and Massimo Buscema.
\newblock Introduction to artificial neural networks.
\newblock {\em European Journal of Gastroenterology \& Hepatology}, 19(12):1046–1054, December 2007.

\bibitem{Hnon1976}
M.~Hénon.
\newblock A two-dimensional mapping with a strange attractor.
\newblock {\em Communications in Mathematical Physics}, 50(1):69–77, February 1976.

\bibitem{Ibrahim2020}
Saleh Ibrahim and Ayman Alharbi.
\newblock Efficient image encryption scheme using henon map, dynamic s-boxes and elliptic curve cryptography.
\newblock {\em IEEE Access}, 8:194289–194302, 2020.

\bibitem{Jrgensen2008}
S.E. Jørgensen.
\newblock {\em Chaos}, page 550–551.
\newblock Elsevier, 2008.

\bibitem{Lai2011}
Ying-Cheng Lai and Tamás Tél.
\newblock {\em Transient Chaos}.
\newblock Springer New York, 2011.

\bibitem{Lai1994}
Ying-Cheng Lai and Raimond~L. Winslow.
\newblock Extreme sensitive dependence on parameters and initial conditions in spatio-temporal chaotic dynamical systems.
\newblock {\em Physica D: Nonlinear Phenomena}, 74(3–4):353–371, July 1994.

\bibitem{Lellep2020}
Martin Lellep, Jonathan Prexl, Moritz Linkmann, and Bruno Eckhardt.
\newblock Using machine learning to predict extreme events in the hénon map.
\newblock {\em Chaos: An Interdisciplinary Journal of Nonlinear Science}, 30(1), January 2020.

\bibitem{Lim}
C.P. Lim, S.C. Woo, A.S. Loh, and R.~Osman.
\newblock Speech recognition using artificial neural networks.
\newblock In {\em Proceedings of the First International Conference on Web Information Systems Engineering}, WISE-00. IEEE Comput. Soc, 2000.

\bibitem{Marotto1979}
Frederick~R. Marotto.
\newblock Chaotic behavior in the henon mapping.
\newblock {\em Communications in Mathematical Physics}, 68(2):187–194, June 1979.

\bibitem{Mishra2018}
Kapil Mishra and Ravi Saharan.
\newblock {\em A Fast Image Encryption Technique Using Henon Chaotic Map}, page 329–339.
\newblock Springer Singapore, December 2018.

\bibitem{Plykin1995}
R.~V. Plykin, E.~A. Sataev, and S.~V. Shlyachkov.
\newblock {\em Strange Attractors}, page 93–139.
\newblock Springer Berlin Heidelberg, 1995.

\bibitem{Pradeepkumar2016}
Dadabada Pradeepkumar and Vadlamani Ravi.
\newblock Forex rate prediction using chaos and quantile regression random forest.
\newblock In {\em 2016 3rd International Conference on Recent Advances in Information Technology (RAIT)}. IEEE, March 2016.

\bibitem{Pratiwi2020}
Heny Pratiwi, Agus~Perdana Windarto, S.~Susliansyah, Ririn~Restu Aria, Susi Susilowati, Luci~Kanti Rahayu, Yuni Fitriani, Agustiena Merdekawati, and Indra~Riyana Rahadjeng.
\newblock Sigmoid activation function in selecting the best model of artificial neural networks.
\newblock {\em Journal of Physics: Conference Series}, 1471(1):012010, February 2020.

\bibitem{Rasamoelina2020}
Andrinandrasana~David Rasamoelina, Fouzia Adjailia, and Peter Sincak.
\newblock A review of activation function for artificial neural network.
\newblock In {\em 2020 IEEE 18th World Symposium on Applied Machine Intelligence and Informatics (SAMI)}. IEEE, January 2020.

\bibitem{RouvasNicolis2009}
Catherine Rouvas-Nicolis and Gregoire Nicolis.
\newblock Butterfly effect.
\newblock {\em Scholarpedia}, 4(5):1720, 2009.

\bibitem{Schulz2012}
Hannes Schulz and Sven Behnke.
\newblock Deep learning: Layer-wise learning of feature hierarchies.
\newblock {\em KI - K\"{u}nstliche Intelligenz}, 26(4):357–363, May 2012.

\bibitem{Sherstinsky2020}
Alex Sherstinsky.
\newblock Fundamentals of recurrent neural network (rnn) and long short-term memory (lstm) network.
\newblock {\em Physica D: Nonlinear Phenomena}, 404:132306, March 2020.

\bibitem{Sukirman2014}
Edi Sukirman, Suryadi MT, and M.~Agus Mubarak.
\newblock The implementation of henon map algorithm for digital image encryption.
\newblock {\em TELKOMNIKA (Telecommunication Computing Electronics and Control)}, 12(3):651, September 2014.

\bibitem{Taye2023}
Mohammad~Mustafa Taye.
\newblock Understanding of machine learning with deep learning: Architectures, workflow, applications and future directions.
\newblock {\em Computers}, 12(5):91, April 2023.

\bibitem{Thenmozhi2013}
S.~Thenmozhi and M.~Chandrasekaran.
\newblock A novel technique for image steganography using nonlinear chaotic map.
\newblock In {\em 2013 7th International Conference on Intelligent Systems and Control (ISCO)}. IEEE, January 2013.

\bibitem{Toker2020}
Daniel Toker, Friedrich~T. Sommer, and Mark D’Esposito.
\newblock A simple method for detecting chaos in nature.
\newblock {\em Communications Biology}, 3(1), January 2020.

\bibitem{Yanan2020}
Guo Yanan, Cao Xiaoqun, Liu Bainian, and Peng Kecheng.
\newblock Chaotic time series prediction using lstm with ceemdan.
\newblock {\em Journal of Physics: Conference Series}, 1617(1):012094, August 2020.

\bibitem{Ye2005}
Meiying Ye.
\newblock {\em Controlling Chaotic Systems via Support Vector Machines Without Analytical Model}, page 919–924.
\newblock Springer Berlin Heidelberg, 2005.

\bibitem{Zhang2020}
Fan Zhang and Lauren~J. O’Donnell.
\newblock {\em Support vector regression}, page 123–140.
\newblock Elsevier, 2020.

\bibitem{Zhang2017}
Lei Zhang.
\newblock Multilayer artificial neural network design and architecture optimization for the pattern recognition and prediction of eeg signals based on hénon map chaotic system.
\newblock In {\em 2017 International Conference on Computational Science and Computational Intelligence (CSCI)}. IEEE, December 2017.

\bibitem{Zhang2018}
Zijun Zhang.
\newblock Improved adam optimizer for deep neural networks.
\newblock In {\em 2018 IEEE/ACM 26th International Symposium on Quality of Service (IWQoS)}. IEEE, June 2018.

\bibitem{Zhou2013}
Tianshou Zhou.
\newblock {\em Bifurcation}, page 79–86.
\newblock Springer New York, 2013.

\bibitem{Zich2020}
Catharina Zich, Andrew~J. Quinn, Lydia~C. Mardell, Nick~S. Ward, and Sven Bestmann.
\newblock Dissecting transient burst events.
\newblock {\em Trends in Cognitive Sciences}, 24(10):784–788, October 2020.

\end{thebibliography}

\end{document}